\pdfoutput=1

\documentclass[11pt]{article}

\usepackage[]{emnlp2021}

\usepackage{times}
\usepackage{latexsym}

\usepackage[T1]{fontenc}

\usepackage[utf8]{inputenc}

\usepackage{microtype}

\usepackage{amsmath}
\usepackage{amsthm}
\usepackage{amssymb}
\usepackage{multirow}
\usepackage{xcolor}
\usepackage{lipsum}
\usepackage{comment}
\usepackage{booktabs}
\usepackage{subcaption} 
\usepackage{graphicx}
\usepackage{bm}
\usepackage{nicefrac}
\usepackage{mathtools} 
\usepackage{listings} 

\theoremstyle{definition}

\newcommand{\norm}[1]{\left\lVert#1\right\rVert}

\renewcommand*{\vec}[1]{\ensuremath{\bm{#1}}} 
\newcommand{\mat}[1]{\bm{#1}}
\newcommand{\loss}{\mathcal{L}}
\newcommand*{\param}{\ensuremath{ \bm{ \theta }}}
\DeclareMathOperator*{\argmin}{arg\,min}
\newcommand{\tdot}[3]{\ensuremath{\langle #1, #2, #3 \rangle}}

\title{Adversarial Attacks on Knowledge Graph Embeddings \\ via Instance Attribution Methods}

\author{Peru Bhardwaj$^{1}$ \quad John Kelleher$^{2}$\thanks{\ \ Equal contribution by last authors.} \quad Luca Costabello$^{3*}$ \quad Declan O'Sullivan$^{1*}$  \\ 
	$^1$ ADAPT Centre, Trinity College Dublin, Ireland \\
	$^2$ ADAPT Centre, TU Dublin, Ireland \\
	$^3$ Accenture Labs, Ireland \\
	\texttt{peru.bhardwaj@adaptcentre.ie} 
}

\begin{document}
\maketitle
\begin{abstract}

Despite the widespread use of Knowledge Graph Embeddings (KGE), little is known about the security vulnerabilities that might disrupt their intended behaviour. We study data poisoning attacks against KGE models for link prediction. These attacks craft adversarial additions or deletions at training time to cause model failure at test time. To select adversarial deletions, we propose to use the model-agnostic \emph{instance attribution methods} from Interpretable Machine Learning, which identify the training instances that are most influential to a neural model's predictions on test instances. We use these influential triples as adversarial deletions. We further propose a heuristic method to replace one of the two entities in each influential triple to generate adversarial additions. Our experiments show that the proposed strategies outperform the state-of-art data poisoning attacks on KGE models and improve the MRR degradation due to the attacks by up to 62\% over the baselines. 

\end{abstract}


\section{Introduction}
Knowledge Graph Embeddings (KGE) are the state-of-art models for relational learning on large scale Knowledge Graphs (KG). They drive enterprise products ranging from search engines to social networks to e-commerce \citep{noy2019knowledgegraphs}. However, the analysis of their security vulnerabilities has received little attention. Identifying these vulnerabilities is especially important for high-stake domains like healthcare and finance that employ KGE models to make critical decisions \citep{hogan2020knowledgegraphs, bendsten2019astrazeneca}. 
We study the security vulnerabilities of KGE models through data poisoning attacks \citep{biggio2018wild,joseph_nelson_rubinstein_tygar_2019} that aim to degrade the predictive performance of learned KGE models by adding or removing triples to the input training graph. 

\captionsetup[figure]{font=small}
\begin{figure}[]
    \centering
    \begin{subfigure}[htb]{1\columnwidth}
        \includegraphics[width=1\columnwidth]{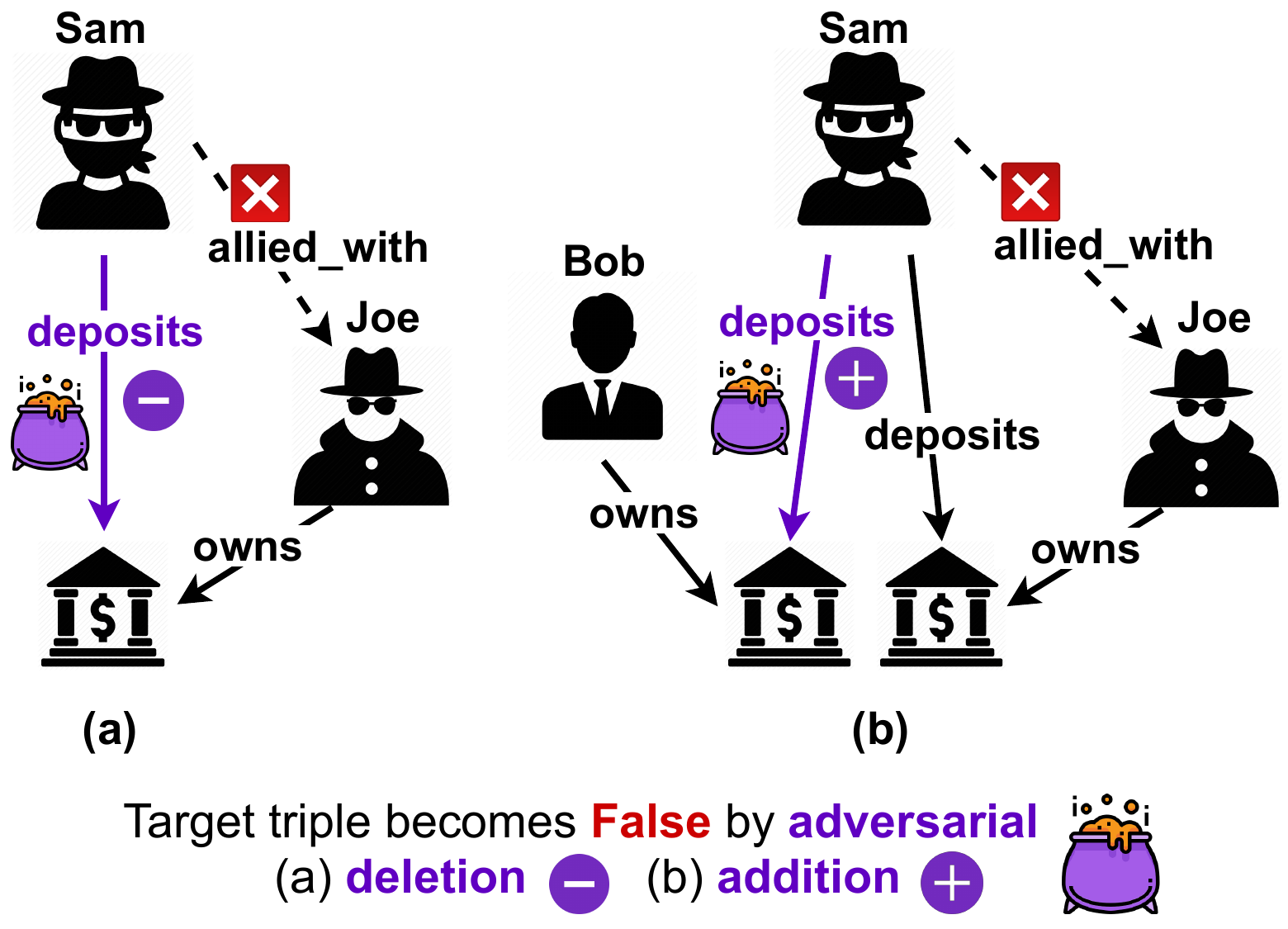}
    \end{subfigure}
    
    \caption{Adversarial attacks against KGE models for fraud detection. The knowledge graph consists of two types of entities - Person and BankAccount. The missing target triple to predict is $(Sam, \mathtt{allied\_with}, Joe)$. Original KGE model predicts this triple as True. But a malicious attacker uses the instance attribution methods to either (a) delete an adversarial triple or (b) add an adversarial triple. Now, the KGE model predicts the missing target triple as False.
    }
    \label{fig:example}
\end{figure}

Designing data poisoning attacks against KGE models poses two main challenges. 
First, to select adversarial deletions or additions, we need to measure the impact of a candidate perturbation on the model's predictions. But the naive approach of re-training a new KGE model for each candidate perturbation is computationally prohibitive. 
Second, while the search space for adversarial \emph{deletions} is limited to existing triples in the KG, it is computationally intractable to enumerate through all candidate adversarial \emph{additions}. 
Furthermore, attack strategies proposed against models for other graph modalities \citep{xu2020advgraphsurvey} do not scale to KGE models; as they would require gradients with respect to a dense adjacency tensor of the KG.

In this work, we propose to use the model-agnostic \emph{instance attribution methods} from Interpretable Machine Learning \citep{molnar2019interpretablemlbook} to select adversarial deletions and additions against KGE models. Instance attribution methods identify the training instances that are \emph{influential} to model predictions, that is, deleting the instances from the training data would considerably change the model parameters or predictions. These methods are widely used to generate post-hoc example-based explanations for deep neural networks on images \citep{koh2017understanding, hanawa2021evaluationsimilaritymetrics, charpiat2019inputsimilarity} and text \citep{han2020influencenlp, han2020influenceveiledtoxicity, pezeshkpour2021instanceattribution}. Since the KGE models have relatively shallow neural architectures and the instance attribution metrics are independent of the black-box models and the input domain, they are a promising approach to estimate the influence of training triples on the KGE model predictions. Yet, despite their promise, they have not been used on KGE models so far. We use the instance attribution methods to address the challenge of measuring the impact of a candidate adversarial deletion on the model predictions.

We focus on the adversarial goal of \emph{degrading} the KGE model prediction on a given \emph{target triple}. To achieve this goal, we use three types of instance attribution methods - Instance Similarity that compares the feature representations of target and training triples \citep{hanawa2021evaluationsimilaritymetrics, charpiat2019inputsimilarity}; Gradient Similarity that compares the gradients of model's loss function due to target and training triples \citep{hanawa2021evaluationsimilaritymetrics,charpiat2019inputsimilarity}; and Influence Function \citep{koh2017understanding} which is a principled approach from the robust statistics to estimate the effect of removing a training triple on the KGE model's predictions. 

Using these metrics, we select the most influential training triple for adversarial deletion. Using the influential triple, we further select adversarial addition by replacing one of the two entities of the influential triple with the most dissimilar entity in the embedding space. The intuition behind this step is to add a triple that would reduce the influence of the influential triple. This solution also overcomes the scalability challenge for adversarial additions by comparing only the entity embeddings to select the replacement. Figure \ref{fig:example} shows an example of the proposed adversarial deletions and additions against KGE models for fraud detection.

We evaluate the proposed attacks for four KGE models - DistMult, ComplEx, ConvE and TransE on two benchmark datasets - WN18RR and FB15k-237. Our results show that instance attribution metrics achieve significantly better performance than \emph{all} state-of-art attacks for both adversarial additions and deletions on three out of four models; and better or equivalent performance on one model. We find that even simple metrics based on instance similarity outperform the state-of-the-art poisoning attacks and are as effective as the computationally expensive Influence Function.

Thus, the main contribution of our research is a collection of effective adversarial deletion and addition strategies based on instance attribution methods against KGE models. 


\section{Knowledge Graph Embeddings}

A Knowledge Graph (KG), is a set of triples \(
 \mathcal{T} \subseteq  \mathcal{E} \times \mathcal{R} \times \mathcal{E}
  \) where each triple encodes the relationship $\mathtt{r}$ as a typed link between the subject entity $s$ and the object entity $o$, i.e. \(
 \mathcal{T} \coloneqq \{t \coloneqq (s,\mathtt{r},o) \, |\,  s,o \in \mathcal{E} \, and \, \mathtt{r} \in \mathcal{R} \}
  \).
  Here, \( \mathcal{E} \) is the set of entities and \( \mathcal{R} \) is the set of relations in the knowledge graph.
  Large scale KGs are often curated automatically from the user content or from the Web and thus are incomplete in practice. 
  To predict the missing links in a KG, the state-of-art method is to learn low dimensional feature vectors for entities and relations in the graph and use them to score the links.
  These feature vectors are called Knowledge Graph Embeddings (KGE) and denoted as \(
 \param \coloneqq \{ \mat{E}, \mat{R}\}
  \) where \( \mat{E} \in \mathbb{R}^{k}\) is the embedding matrix for entities, \( \mat{R} \in \mathbb{R}^{k}\) is the embedding matrix for relations and $k$ is the embedding dimension.

\paragraph{Scoring Functions:}
KGE models differ from each other by their scoring functions \(f : \mathcal{T} \rightarrow \mathbb{R} \) which combine the subject, relation and object embeddings to assign a score to the triple, i.e. \(f_{t} \coloneqq f(\vec{e}_s, \vec{e}_\mathtt{r}, \vec{e}_o) \) where \(\vec{e}_s, \vec{e}_o \in \mat{E}\) and \(\vec{e}_\mathtt{r} \in \mat{R}\). 
Table \ref{tab:scoring_functions} shows the different scoring functions of KGE models used in this research.

These scoring functions are used to categorize the models as additive or multiplicative \cite{chandrahas2018towards}.
\emph{Additive} models apply relation-specific translation from the subject embedding to the object embedding. The scoring function for such models is expressed as $f_{t} = -\norm{\mat{M}_{\mathtt{r}}^1(\vec{e}_s) + \vec{e}_\mathtt{r} - \mat{M}_{\mathtt{r}}^2(\vec{e}_o)}$ where $\mat{M}_\mathtt{r} \in \mathbb{R}^{k \times k}$ is a projection matrix from entity space to relation space. An example of additive models is TransE where $\mat{M}_{\mathtt{r}}^1 = \mat{M}_{\mathtt{r}}^2 = \mat{I}$. 

On the other hand, \emph{multiplicative} models score triples through multiplicative interactions between the subject, relation and object embeddings. The scoring function for these models is expressed as $f_{t} = \vec{e}_{\mathtt{r}}^\top \mathcal{F}(\vec{e}_s, \vec{e}_o)$ where the function $\mathcal{F}$ measures the compatibility between the subject and object embeddings and varies across different models within this family. DistMult, ComplEx and ConvE are examples of multiplicative models.

\begin{table}[]
    \centering
    \small
    \begin{tabular}{c c c}
    \toprule
    
     \textbf{Model} & \multicolumn{1}{p{2.5cm}}{\centering \bf Scoring Function} & \multicolumn{1}{p{2.5cm}}{\centering \bf Feature Vectors}  \\
     \midrule
       DistMult  &  $ \tdot{\vec{e}_s}{\vec{e}_{\mathtt{r}}}{\vec{e}_o}$  & $ \vec{e}_s \circ \vec{e}_{\mathtt{r}} \circ \vec{e}_o$  \\ [2pt]
       ComplEx  &  $ \Re(\tdot{\vec{e}_s}{\vec{e}_{\mathtt{r}}}{\overline{\vec{e}_o}})$   &   $ \Re(\vec{e}_s \circ \vec{e}_{\mathtt{r}} \circ \overline{\vec{e}_o})$ \\ [2pt]
       ConvE    &  $ \langle (\vec{e}_s * \vec{e}_{\mathtt{r}}), \vec{e}_o \rangle$  &  $ (\vec{e}_s * \vec{e}_{\mathtt{r}}) \circ \vec{e}_o $ \\ [2pt]
       TransE   &  $- \norm{\vec{e}_s + \vec{e}_{\mathtt{r}} - \vec{e}_o}_p$  &  $- (\vec{e}_s + \vec{e}_{\mathtt{r}} - \vec{e}_o)$ \\ [2pt]
       \bottomrule
    \end{tabular}
    \caption{Scoring functions \(f_{s\mathtt{r}o}\) and the proposed Triple Feature Vectors \(\vec{f}_{s\mathtt{r}o}\) of the KGE models used in this research. For ComplEx, $\vec{e}_s, \vec{e}_\mathtt{r}, \vec{e}_o \in \mathbb{C}^k$; for the remaining models $\vec{e}_s, \vec{e}_\mathtt{r}, \vec{e}_o \in \mathbb{R}^k$. Here, $\langle \cdot \rangle$ denotes the tri-linear dot product; $\circ$ denotes the element-wise Hadamard product; $\overline{\ \cdot\ }$ denotes conjugate for complex vectors; $\norm{\cdot}_p$ denotes l-p norm; $*$ is the neural architecture in ConvE, i.e. \(\vec{e}_s * \vec{e}_\mathtt{r} \coloneqq \sigma(\mathrm{vec}(\sigma([ \overline{\vec{e}_\mathtt{r}}, \overline{\vec{e}_s}] \ast \boldsymbol{\Omega})) \mat{W})\) where $\sigma$ denotes sigmoid activation, $\ast$ denotes 2D convolution; $\overline{\ \cdot\ }$ denotes 2D reshaping of real vectors.}
    \label{tab:scoring_functions}
\end{table}

\paragraph{Training:}
  Since the KGs only contain positive triples; to train the KGE model, synthetic negative samples \(t' \in \mathcal{T}'\) are generated by replacing the subject or object in the positive triples with other entities in $\mathcal{E}$. That is, for each positive triple $t \coloneqq (s,\mathtt{r},o)$, the set of negative samples is $ t' \coloneqq \{(s',\mathtt{r},o) \cup (s,\mathtt{r},o')\}$. The training objective is to learn the embeddings that score positive triples existing in the KG higher than the negative triples generated synthetically. To achieve this, a triple-wise loss function \(
  \loss(t,\param) \coloneqq \ell(t,\param) + \sum_{t' \in \mathcal{T}'} \ell(t', \param)
  \) is minimized.
  Thus, the optimal parameters \(\widehat \param\) learned by the model are defined by \(
  \widehat{\param} \coloneqq \argmin_{\param} \sum_{t \in \mathcal{T}} \loss(t, \param)
  \). 
Further details on KGE loss functions and negative sampling strategies are available in \citet{ruffinelli2020olddognewtricks}.

\paragraph{Missing Link Prediction: }
Given the learned embeddings \(\param\), missing triples in the knowledge graph are predicted by an entity ranking evaluation protocol. 
Similar to the training process, subject-side negatives $t_s'={(s',\mathtt{r},o)}$ and object-side negatives $t_o'={(s,\mathtt{r},o')}$ are sampled for each test triple $t=(s,\mathtt{r},o)$ to be predicted. Of these negatives, the triples already existing in the training, validation or test set are filtered out \citep{bordes2013transe}.  The test triple is then ranked against the remaining negatives based on the scores predicted by the KGE model. 
The state-of-art evaluation metrics reported over the entire set are (i) \emph{MR}: mean of the ranks, (ii) \emph{MRR}: mean of the reciprocals of ranks and (iii) \emph{Hits@n}: number of triples ranked in top-n. 


\section{Poisoning Knowledge Graph Embeddings via Instance Attribution}
\label{sec:poisoning}
We consider an adversarial attacker that aims to degrade the KGE model's predictive performance on a set of missing triples that have been ranked highly plausible by the model. 
We denote these \emph{target triples} as \(\mathcal{Z} \coloneqq \{z \coloneqq (z_s, z_\mathtt{r}, z_o)\}\). Since the predicted ranks are based on the predicted scores; to reduce the predicted rank of a target triple, we craft perturbations to the training data that aim to reduce the predicted score of the target triple.

\paragraph{Threat Model:}We use the same threat model as the state-of-art poisoning attacks on KGE models \citep{pezeshkpour2019criage, zhang2019kgeattack}. We focus on the white-box attack setting where the attacker has full knowledge of the victim model architecture and access to the learned embeddings. However, they cannot perturb the architecture or the embeddings directly; but only through perturbations in the training data. 
We study both \emph{adversarial additions} and \emph{adversarial deletions}. In both settings, the attacker is restricted to making only one edit in the neighbourhood of the target triple. The neighbourhood of the target triple $z \coloneqq (z_s,z_\mathtt{r},z_o)$ is the set of triples that have the same subject or the same object as the target triple, i.e. $\mathcal{X}:= \{x \coloneqq (x_s, x_\mathtt{r}, x_o)\, |\, x_s \in \{z_s, z_o\} \vee x_o \in \{z_s, z_o\}\}$.

\subsection{Instance Attribution Methods}
For adversarial deletions, we want to identify the training triples that have influenced the KGE model's prediction on the target triple. Deleting these influential triples from the training set will likely degrade the prediction on the target triple. 
Thus, we define an \emph{influence score} \(\phi (z,x) : \mathcal{T} \times \mathcal{T} \rightarrow \mathbb{R} \) for the pairs of triples \((z,x) \in \mathcal{T} \times \mathcal{T}\) which indicates the influence of training triple $x$ on the prediction of target triple $z$.
Larger values of the influence score \(\phi (z,x)\) indicate that removing $x$ from the training data would cause larger reduction in the predicted score on $z$. 

Trivially, we can compute the influence score for a training triple by removing the triple and re-training the KGE model. However, this is a prohibitively expensive step that requires re-training a new KGE model for every candidate influential triple.
Thus, we use the following instance-attribution methods from Interpretable Machine Learning \citep{molnar2019interpretablemlbook} to estimate the influence score \(\phi (z,x)\) without re-training the model. 
\subsubsection{Instance Similarity}
We estimate the influence of training triple $x$ on the prediction of target triple $z$ based on the similarity of their feature representations. The intuition behind these metrics is to identify the training triples that a KGE model has learnt to be similar to the target triple and thus (might) have influenced the model's prediction on the target triple.

Computing this similarity between triples requires feature vector representations for the triples. We note that while the standard KGE scoring functions assign a scalar score to the triples, this scalar value is obtained by reducing over the embedding dimension. For example, in the tri-linear dot product for DistMult, the embeddings of subject, relation and object are multiplied element-wise and then the scalar score for the triple is obtained by summing over the embedding dimension, i.e. $f_t \coloneqq \langle \vec{e}_s, \vec{e}_\mathtt{r}, \vec{e}_o\rangle \coloneqq \sum_{i=1}^k \vec{e}_{s_i} \vec{e}_{\mathtt{r}_i} \vec{e}_{o_i}$ where $k$ is the embedding dimension. 

Thus, to obtain feature vector representations for the triples \(\vec{f}_t : \mathcal{E} \times \mathcal{R} \times \mathcal{E} \rightarrow \mathbb{R}^k \), we use the state-of-art KGE scoring functions without reduction over the embedding dimension. For the DistMult model, the triple feature vector is \(\vec{f} \coloneqq \vec{e}_s \circ \vec{e}_\mathtt{r} \circ \vec{e}_o\) where $\circ$ is the Hadamard (element-wise) product. Table \ref{tab:scoring_functions} shows the feature vector scores for different KGE models used in this research.

Given the feature vectors for target triples $\vec{f}(z)$ and training triples $\vec{f}(x)$, we follow \citet{hanawa2021evaluationsimilaritymetrics} and define the following metrics.

\textbf{Dot Metric:} This metric computes the similarity between target and training instances as the dot product of their feature vectors.
That is, \(
 \phi_{dot} (z,x) \coloneqq \langle \vec{f}(z), \vec{f}(x) \rangle
\)

\textbf{$\bm{\ell_2}$ Metric: } This metric computes similarity as the negative Euclidean distance between the feature vectors of target instance and test instance. 
That is,
\(
\phi_{\ell_2} (z,x) \coloneqq - \norm{\vec{f}(z) - \vec{f}(x)}_2
\)

\textbf{Cosine Metric:} This metric computes similarity as the dot product between $\bm{\ell_2}$ normalized feature vectors of target and test instance, i.e. it ignores the magnitude of the vectors and only relies on the angle between them.
That is, \break
\(
\phi_{cos} (z,x) \coloneqq \cos{(\vec{f}(z), \vec{f}(x))}
\)

Here, we denote the dot product for two vectors $\vec{a}$ and $\vec{b}$ as $\langle \vec{a}, \vec{b} \rangle \coloneqq \sum_{i=1}^p a_i b_i$; the $\bm{\ell_2}$ norm of a vector as $\norm{\vec{a}}_2 \coloneqq \sqrt{\langle \vec{a}, \vec{a} \rangle}$; and the cos similarity between vectors $\vec{a}$ and $\vec{b}$ as \break $\cos(\vec{a}, \vec{b}) \coloneqq \nicefrac{\langle \vec{a}, \vec{b} \rangle}{\norm{\vec{a}}_2 \norm{\vec{b}}_2}$.

\subsubsection{Gradient Similarity}
We represent the gradient of the loss for triple $z$ w.r.t. model parameters as \(\vec{g}(z, \widehat{\param}) := \nabla_{\param}{\loss (z,\widehat{\param)}} \). Gradient similarity metrics compute similarity between the gradients due to target triple $z$ and the gradients due to training triple $x$. The intuition is to assign higher influence to training triples that have similar effect on the model's parameters as the target triple; and are therefore likely to impact the prediction on target triple \citep{charpiat2019inputsimilarity}. 
Thus, using the same similarity functions as Instance Similarity metrics, we define the following three metrics for gradient similarity - Gradient Dot (GD), Gradient $\bm{\ell_2}$ (GL) and Gradient Cosine (GC).

\paragraph{GD(dot): } \(\phi_{\mathrm{GD}} (z,x) \coloneqq \langle \, \vec{g}(z, \widehat{\param})\, , \, \vec{g}(x, \widehat{\param}) \, \rangle \)

\paragraph{GL ($\bm{\ell_2}$):} \(\phi_{\mathrm{GL}} (z,x) \coloneqq - \norm{\vec{g}(z, \widehat{\param}) - \vec{g}(x, \widehat{\param})}_2 \)

\paragraph{GC(cos):} \(\phi_{\mathrm{GC}} (z,x) \coloneqq \cos{( \, \vec{g}(z, \widehat{\param}) \, , \, \vec{g}(x, \widehat{\param}) \, )} \)

\subsubsection{Influence Functions}
Influence Functions (IF) is a classic technique from robust statistics and was introduced to explain the predictions of black-box models in \citet{koh2017understanding}. To estimate the effect of a training point on a model's predictions, it first approximates the effect of removing the training point on the learned model parameters. To do this, it performs a first order Taylor expansion around the learned parameters $\widehat{\param}$ at the optimality conditions. 

Following the derivation in \citet{koh2017understanding}, the the effect of removing the training triple $x$ on $\widehat{\param}$ is given by \(
\nicefrac{d\widehat{\param}}{d\epsilon_i} = \mat{H}_{\widehat{\param}}^{-1} \ \vec{g}(x, \widehat{\param})
\). Here, $\mat{H}_{\widehat{\param}}$ denotes the Hessian of the loss function \(
\mat{H}_{\widehat{\param}} \coloneqq \nicefrac{1}{n} \sum_{t \in \mathcal{T}} \nabla_{\param}^{2}{\loss (t, \widehat{\param})}
\). 
Using the chain rule then, we approximate the influence of removing $x$ on the model's prediction at $z$ as $\langle \vec{g} (z,\widehat{\param}) \ ,  \nicefrac{d\widehat{\param}}{d\epsilon_i} \rangle$. Thus, the influence score using IF is defined as:

\textbf{IF: } \, \(\phi_{\mathrm{IF}} (z,x) := \langle \, \vec{g}(z, \widehat{\param}) \, , \, \mat{H}_{\widehat{\param}}^{-1} \vec{g}(x, \widehat{\param}) \, \rangle \)

Computing the IF for KGE models poses two challenges - (i) storing and inverting the Hessian matrix is computationally too expensive for a large number of parameters; (ii) the Hessian is not guaranteed to be positive definite and thus, invertible because KGE models are non-convex models. To address both these challenges, we follow the guidelines in \citet{koh2017understanding}. Instead of computing the exact Hessian matrix, we estimate the Hessian-vector product (HVP) with target triple's gradient. That is, for every target triple $z$, we pre-compute the value \(
\mat{H}_{\widehat{\param}}^{-1} \ \vec{g} (z, \widehat{\param})
\). Then, for each neighbourhood triple $x$ in the training set, we compute $\phi_{\mathrm{IF}} (z,x)$ using the pre-computed HVP. Furthermore, we use the stochastic estimator LiSSA \citep{agarwal2017lissa} that computes the HVP in linear time using samples from training data.
For the second issue of non-convexity, we add a "damping" term to the Hessian so that it is positive definite and invertible. This term is a hyperparameter that is tuned to ensure that all eigenvalues of the Hessian matrix are positive, i.e. the Hessian matrix is positive definite. Further discussion on the validity of Influence Functions for non-convex settings is available in \citet{koh2017understanding}.

\subsection{Adversarial Additions}
In this attack setting, the adversarial attacker can only \emph{add} triples to the neighbourhood of target triple.
Using the Instance Attribution metrics above, we select the training triple $x := (x_s, x_\mathtt{r}, x_o)$ in the neighbourhood of the target triple $z := (z_s, z_\mathtt{r}, z_o)$ that is most influential to the prediction of $z$. For brevity, lets assume $x_s = z_s$, i.e. the influential and target triples have the same subject. To generate adversarial addition using the influential triple, we propose to replace $x_o$ with the most dissimilar entity $x_{o'}$. Since the adversarial triple $x' := (x_s, x_\mathtt{r}, x_{o'})$ has the same subject and relation as the influential triple but a different object, it should reduce the influence of the influential triple on the target triple's prediction. This in turn should degrade the model prediction on target triple. 
For multiplicative models, we select the dissimilar entity $x_{o'}$ using the cosine similarity between $x_o$ and the entities $\mathcal{E}$. For additive models, we use the $\bm{\ell_2}$ similarity between $x_o$ and the entities $\mathcal{E}$.


\section{Evaluation}
\label{sec:evaluation}
We evaluate the effectiveness of the proposed attack strategies in degrading the KGE model's predictions on target triples at test time.
We follow the state-of-art protocol to evaluate poisoning attacks \citep{xu2020advgraphsurvey} - we train a victim KGE model on the original dataset; generate adversarial deletions or additions using one of the attacks; perturb the original dataset; and train a new KGE model on the perturbed dataset. The hyperparameters for victim and poisoned KGE models are same.

We evaluate our attacks on four state-of-art KGE models - DistMult, ComplEx, ConvE and TransE on two publicly available\footnote{https://github.com/TimDettmers/ConvE} benchmark datasets - WN18RR and FB15k-237. 
To be able to evaluate the effectiveness of attacks in \emph{degrading} the predictive performance, we select a subset of the benchmark test triples that has been ranked \emph{highest} (ranks=1) by the victim KGE model. From this subset, we randomly sample 100 triples as the \emph{target triples}. This is to avoid the expensive Hessian inverse estimation in the IF metric for a large number of target triples (for each target triple, this estimation requires one training epoch).

The source code implementation of our experiments is available at \url{https://github.com/PeruBhardwaj/AttributionAttack}.

\paragraph{Baselines: }
We evaluate our attacks against baseline methods based on random edits and the state-of-art poisoning attacks. \emph{Random\_n} adds or removes a random triple from the neighbourhood of the target triple. \emph{Random\_g} adds or removes a random triple globally and is not restricted to the target's neighbourhood. 
\emph{Direct-Del} and \emph{Direct-Add} are the adversarial deletion and addition attacks proposed in \citet{zhang2019kgeattack}. \emph{CRIAGE} is the poisoning attack from \citet{pezeshkpour2019criage} and is a baseline for both deletions and additions. 
\emph{GR} (Gradient Rollback) \citep{lawrence2021gradientrollback} uses influence estimation to provide post-hoc explanations for KGE models and can also be used to generate adversarial deletions. Thus, we include this method as a baseline for adversarial deletions. 

The attack evaluations in \citet{zhang2019kgeattack, pezeshkpour2019criage, lawrence2021gradientrollback} differ with respect to the definition of their \emph{neighbourhood}. Thus, to ensure fair evaluation, we implement all methods with the same neighbourhood - triples that are linked to the subject or object of the target triple (Section \ref{sec:poisoning}). We use the publicly available implementations for CRIAGE\footnote{https://github.com/pouyapez/criage} and Gradient Rollback\footnote{https://github.com/carolinlawrence/gradient-rollback} and implement Direct-Del and Direct-Add ourselves. Further details on datasets, implementation of KGE models, baselines and computing resources is available in Appendix \ref{apx:dataset_details} and \ref{apx:training_details}.

\begin{table*}
\centering
\footnotesize
\setlength{\tabcolsep}{3.5pt}
\begin{tabular}{c  l  ll  ll   ll  ll }

\toprule
    
     & & \multicolumn{2}{c}{\textbf{DistMult}} & \multicolumn{2}{c}{\textbf{ComplEx}} & \multicolumn{2}{c}{\textbf{ConvE}} & \multicolumn{2}{c}{\textbf{TransE}} \\
   \cmidrule(lr){3-4}  \cmidrule(lr){5-6}  \cmidrule(lr){7-8} \cmidrule(lr){9-10} 
    & & \textbf{MRR}   & \textbf{Hits@1}  & \textbf{MRR}   & \textbf{Hits@1} & \textbf{MRR}   & \textbf{Hits@1} & \textbf{MRR}   & \textbf{Hits@1} \\
\midrule
    
     \textbf{Original}  &    & 1.00            &  1.00     &  1.00             & 1.00      &  1.00          &   1.00    & 1.00           &  1.00 \\
\midrule
    \multirow{6}{*}{\shortstack[l]{\textbf{Baseline} \\ \textbf{Attacks}}}
    & Random\_n              & 0.87 (-13\%) &  0.82     & 0.85 (-15\%)   & 0.80        & 0.82 (-18\%)         & 0.79        &    0.82 (-18\%) &  0.70   \\
    & Random\_g              & 0.97         &  0.95     & 0.96           & 0.93        &  0.99                & 0.98       &   0.93           &  0.87   \\
\cline{2-10} \\[-7pt]
    & Direct-Del             & 0.88         &  0.77     &  0.86 (-14\%)  &  0.77        &  0.71 (-29\%)        & 0.64        &  0.54 \textbf{(-46\%)}  & 0.37    \\
    & CRIAGE                 & 0.73 (-27\%) &  0.66     &   -             &  -          &  Er                  & Er        &   -          &  -   \\
    & GR                      & 0.95        &  0.90     &  0.93           &  0.86       & 0.95                 & 0.91      &   0.84           &  0.77  \\
\midrule
    \multirow{9}{*}{\shortstack[l]{\textbf{Proposed} \\ \textbf{Attacks}}}
    & Dot Metric               & 0.89                  &  0.82     & 0.85           & 0.79         & 0.84 (-16\%)      & 0.80       &  0.77     & 0.60   \\
    & $\bm{\ell_2}$ Metric     & \textbf{0.25 (-75\%)} &  0.16      & 0.29 (-71\%)  & 0.20         &  0.88      &  0.78       &   0.62    & 0.50   \\
    & Cos Metric               & \textbf{0.25 (-75\%)} & 0.16      & 0.29 (-71\%)   &  0.20        & 0.87        & 0.76      &  0.56 (-44\%)    & 0.40   \\
\cline{2-10} \\[-7pt] 
    & GD (dot)                  &   0.28 (-72\%) &  0.19        &  0.29                    & 0.21        &  0.25          & 0.21        &  0.71 (-29\%)   & 0.57    \\
    & GL ($\bm{\ell_2}$)        &  0.30          &  0.20        &  0.28 \textbf{(-72\%)}   &  0.19       &  0.17 \textbf{(-83\%)}      & 0.12        &   0.72     & 0.60    \\
    & GC (cos)                  &  0.29          &  0.19        &  0.29                    & 0.21        & 0.20   & 0.16        & 0.71 (-29\%)    & 0.57   \\
\cline{2-10} \\[-7pt]
    
    & IF                       &  0.28 (-72\%)   & 0.19        & 0.29 (-71\%)    & 0.20        & 0.22 (-78\%)          & 0.17     &  0.71 (-29\%)     &  0.57  \\
     
\bottomrule    

\end{tabular}
\caption{\small Reduction in MRR and Hits@1 due to \textbf{adversarial deletions on target triples in WN18RR}. Lower values indicate better results; best results for each model are in bold. First block of rows are the baseline attacks with random edits; second block is state-of-art attacks; remaining are the proposed attacks. For each block, we report the \emph{best} reduction in percentage relative to the original MRR; computed as $(poisoned - original)/original*100$. }
\label{tab:deletion_WN18RR}
\end{table*}

\paragraph{Results:}
For WN18RR and FB15k-237 respectively, Tables \ref{tab:deletion_WN18RR} and \ref{tab:deletion_FB15k-237} show the degradation in MRR and Hits@1 due to adversarial deletions; and Tables \ref{tab:addition_WN18RR} and \ref{tab:addition_FB15k-237} due to adversarial additions for state-of-art KGE models. 
Below we discuss different patterns in these results. We also discuss runtime efficiency of the attack methods in Appendix \ref{apx:runtime_analysis}.

\subsection{Comparison with Baselines}
\label{sec:eval_baseline}
We observe that the proposed strategies for adversarial deletions and adversarial additions successfully degrade the predictive performance of KGE models. 
On the other hand, the state-of-art attacks are ineffective or only partially effective. Adversarial deletions from Gradient Rollback perform similar to random baselines; likely because this method estimates the influence of a training triple as the sum of its gradients over the training process. In this way, it does not account for the target triple in the influence estimation. The method is also likely to be effective only for a KGE model that is trained with a batch size of 1 because it needs to track the gradient updates for each triple.

The CRIAGE baseline is only applicable to DistMult and ConvE. But we found that the method ran into \texttt{\small numpy.linalg.LinAlgError: Singular matrix} error for ConvE; because the Hessian matrix computed from the victim model embeddings was non-invertible\footnote{This issue might be resolved by changing the hyperparameters of the victim KGE model so that the Hessian matrix from the victim embeddings is invertible. But there is no strategic way to make such changes.}. For adversarial deletions on DistMult, the baseline works better than random edits but not the proposed attacks \footnote{Since the influence estimation in CRIAGE uses BCE loss, we also compare for DistMult trained with BCE in Appendix \ref{apx:criage_bce}, but the results are similar.}.
It is also ineffective against adversarial additions. 

We see that Direct-Del is effective on TransE, but not on multiplicative models. 
This is likely because it estimates the influence of a candidate triple as the \emph{difference} in the triple's score when the neighbour entity embedding is perturbed. The additive nature of this influence score might make it more suitable for additive models. We also see that Direct-Add works similar to random additions, likely because it uses random down-sampling.

The proposed attacks based on instance attribution methods consistently outperform random baselines for adversarial additions and deletions. One exception to this pattern are adversarial additions against TransE on WN18RR. In this case, no influence metric performs better than random neighbourhood edits, though they are all effective for adversarial deletions. One possible reason is that the TransE model is designed to learn hierarchical relations like $\mathtt{\_has\_part}$. We found that the target triples ranked highest by the model have such hierarchical relations; and the influential triple for them has the same relation. That is, the triple $(s_1, \mathtt{\_has\_part}, s)$ is the influential triple for $(s, \mathtt{\_has\_part}, o)$. Removing this influential triple breaks the hierarchical link between $s_1$ and $s$; and degrades TransE predictions on the target. But adding the triple $(s_2, \mathtt{\_has\_part}, s)$ still preserves the hierarchical structure which TransE can use to score the target correctly. We provide more examples of such relations in Appendix \ref{apx:wn18rr_transe}.

\subsection{Comparison across Influence Metrics }
\label{sec:eval_if}
We see that the IF and Gradient Similarity metrics show similar degradation in predictive performance. This indicates that the computationally expensive Hessian inverse in the IF can be avoided and simpler metrics can identify influential triples with comparable effectiveness. 
Furthermore, cos and $\bm{\ell_2}$ based Instance Similarity metrics outperform all other methods for adversarial deletions on DistMult, ComplEx and TransE. This effectiveness of naive metrics indicates the high vulnerability of shallow KGE architectures to data poisoning attacks in practice.
In contrast to this, the Input Similarity metrics are less effective in poisoning ConvE, especially significantly on WN18RR. This is likely because the triple feature vectors for ConvE are based on the output from a deeper neural architecture than the Embedding layer alone.
Within Instance Similarity metrics, we see that the dot metric is not as effective as others. This could be because the dot product does not normalize the triple feature vectors. Thus, training triples with large norms are prioritized over relevant influential triples \citep{hanawa2021evaluationsimilaritymetrics}.

\subsection{Comparison of datasets}
\label{sec:eval_data}
We note that the degradation in predictive performance is more significant on WN18RR than on FB15k-237. This is likely due to the sparser graph structure of WN18RR, i.e. there are fewer neighbours per target triple in WN18RR than in FB15k-237 (Appendix \ref{apx:neighbourhood_sparsity}). Thus, the model learns its predictions from few influential triples in WN18RR; and removing only one neighbour significantly degrades the model's predictions on the target triple. 

On the other hand, because of more neighbours in FB15k-237, the model predictions are likely influenced by a \emph{group} of training triples. Such group effect of training instances on model parameters has been studied in \citet{koh2019groupinfluence, basu2020groupinfluence}. We will investigate these methods for \emph{KGE models} on FB15k-237 in the future.

\begin{table*}
\centering
\footnotesize
\setlength{\tabcolsep}{3.5pt}
\begin{tabular}{c  l  ll  ll   ll  ll }

\toprule
    
     & & \multicolumn{2}{c}{\textbf{DistMult}} & \multicolumn{2}{c}{\textbf{ComplEx}} & \multicolumn{2}{c}{\textbf{ConvE}} & \multicolumn{2}{c}{\textbf{TransE}} \\
   \cmidrule(lr){3-4}  \cmidrule(lr){5-6}  \cmidrule(lr){7-8} \cmidrule(lr){9-10} 
    & & \textbf{MRR}   & \textbf{Hits@1}  & \textbf{MRR}   & \textbf{Hits@1} & \textbf{MRR}   & \textbf{Hits@1} & \textbf{MRR}   & \textbf{Hits@1} \\
\midrule
    
     \textbf{Original}  &    & 1.00            &  1.00     &  1.00             & 1.00      &  1.00          &   1.00    & 1.00           &  1.00 \\
\midrule
    \multirow{6}{*}{\shortstack[l]{\textbf{Baseline} \\ \textbf{Attacks}}}
    & Random\_n              &  0.66 (-34\%)  &  0.52      &   0.65 (-35\%)   & 0.51           & 0.62 (-38\%)    &  0.46       &   0.71 (-29\%)   &  0.56      \\
    & Random\_g              &  0.68          &  0.53      &  0.65 (-35\%)    &  0.51         &  0.63            &  0.50       &   0.75           &   0.61     \\
\cline{2-10} \\[-7pt]
    & Direct-Del             & 0.59 (-41\%)     &  0.42     &   0.62 (-38\%)    &  0.47     &   0.57 (-43\%)          &  0.41       &   0.62 \textbf{(-38\%)}          &   0.45      \\
    & CRIAGE                 & 0.62             & 0.47      &   -               &  -        &      Er        &    Er     &       -       &     -    \\
    & GR           & 0.68             & 0.55      &  0.66             &  0.51     &  0.62            & 0.45        &   0.68           &   0.53      \\
\midrule
    \multirow{9}{*}{\shortstack[l]{\textbf{Proposed} \\ \textbf{Attacks}}}
    & Dot Metric               & 0.63                  & 0.47        & 0.64                  & 0.49        & 0.60        & 0.44       &   0.74        &  0.62       \\
    & $\bm{\ell_2}$ Metric     & 0.58                  & 0.41        & 0.56 \textbf{(-44\%)} &  0.40       & 0.53 \textbf{(-47\%)}       & 0.35        &   0.63 (-37\%)      &  0.46        \\
    & Cos Metric               & 0.56 \textbf{(-44\%)}  & 0.39       & 0.57                  & 0.40        & 0.55        & 0.38        &  0.63 (-37\%)       &  0.45        \\
\cline{2-10} \\[-7pt] 
    & GD (dot)           &   0.60          & 0.44         &  0.60         & 0.45        & 0.55 (-45\%)     & 0.37        &  0.65           &   0.49              \\
    & GL ($\bm{\ell_2}$)   &   0.62          &  0.45        &  0.60         &  0.45       & 0.56      & 0.41        &   0.70          &   0.58              \\
    & GC (cos)           &   0.58 (-42\%)  &  0.42        &  0.57 (-43\%) & 0.39        & 0.57      & 0.40        &   0.64 (-36\%)         &  0.48               \\
\cline{2-10} \\[-7pt]
    
    & IF       &  0.60 (-40\%)        &  0.44      &  0.60 (-40\%)   & 0.45         & 0.58 (-42\%)          & 0.43     &  0.66 (-34\%)     &    0.52               \\
    
\bottomrule    

\end{tabular}
\caption{\small Reduction in MRR and Hits@1 due to \textbf{adversarial deletions on target triples in FB15k-237}. Lower values indicate better results. First block of rows are the baseline attacks with random edits; second block is state-of-art attacks; remaining are the proposed attacks. For each block, we report the \emph{best} reduction in percentage relative to the original MRR.}
\label{tab:deletion_FB15k-237}
\end{table*}

\begin{table*}
\centering
\footnotesize
\setlength{\tabcolsep}{3.5pt}
\begin{tabular}{c  l  ll  ll   ll  ll }

\toprule
    
     & & \multicolumn{2}{c}{\textbf{DistMult}} & \multicolumn{2}{c}{\textbf{ComplEx}} & \multicolumn{2}{c}{\textbf{ConvE}} & \multicolumn{2}{c}{\textbf{TransE}} \\
   \cmidrule(lr){3-4}  \cmidrule(lr){5-6}  \cmidrule(lr){7-8} \cmidrule(lr){9-10} 
    & & \textbf{MRR}   & \textbf{Hits@1}  & \textbf{MRR}   & \textbf{Hits@1} & \textbf{MRR}   & \textbf{Hits@1} & \textbf{MRR}   & \textbf{Hits@1} \\
\midrule
    
     \textbf{Original}  &    & 1.00            &  1.00     &  1.00             & 1.00      &  1.00          &   1.00    & 1.00           &  1.00 \\
\midrule
    \multirow{5}{*}{\shortstack[l]{\textbf{Baseline} \\ \textbf{Attacks}}}
    & Random\_n              & 0.99 (-1\%)       & 0.98       & 0.97 (-3\%)    & 0.94           &  0.99 (-1\%)           & 0.98       &   0.76 \textbf{(-24\%)}          &  0.57      \\
    & Random\_g              & 0.99 (-1\%)     & 0.97        &  0.97 (-3\%)    & 0.95          &   0.99 (-1\%)          &  0.98       &   0.93           &  0.87      \\
\cline{2-10} \\[-7pt]
    & Direct-Add             & 0.98 (-2\%)     &  0.96      &  0.95 (-5\%)    &  0.92        &  0.99 (-1\%)           &  0.98       &   0.81 (-19\%)          &   0.67      \\
    & CRIAGE                 & 0.98 (-2\%)     & 0.97       &  -               &  -         &    Er          &  Er                  &     -         &    -     \\
\midrule
    \multirow{9}{*}{\shortstack[l]{\textbf{Proposed} \\ \textbf{Attacks}}}
    & Dot Metric               &  0.97                      & 0.93        & 0.95         & 0.90        &   0.95 (-5\%)    &  0.91      &  0.95         & 0.90        \\
    & $\bm{\ell_2}$ Metric     &  0.89 \textbf{(-11\%)}     & 0.78        &  0.88        &  0.77       &   0.98           &   0.96      &  0.87 (-13\%)       & 0.83         \\
    & Cos Metric               & 0.89 \textbf{(-11\%)}      & 0.78        &  0.87 (-13\%) & 0.77        &   0.99          &   0.98      &  0.87 (-13\%)       &  0.83        \\
\cline{2-10} \\[-7pt] 
    & GD (dot)            &  0.90                     & 0.79         & 0.89                    & 0.79        &  0.92     & 0.85                 &   0.80 (-20\%)         &  0.73               \\
    & GL ($\bm{\ell_2}$)  &  0.89 \textbf{(-11\%)}    & 0.79         & 0.86 \textbf{(-14\%)}   &  0.73       &  0.88 \textbf{(-12\%)} & 0.77    &   0.89          &  0.83               \\
    & GC (cos)           &  0.90                     & 0.80         &  0.87                   &  0.76       &  0.91     &  0.82                &   0.80 (-20\%)         &  0.73               \\
\cline{2-10} \\[-7pt]
    
    & IF           &  0.90 (-10\%)          & 0.79       &  0.89 (-11\%)              &  0.79       &  0.91 (-8.9\%)         & 0.82     &   0.77 (-23\%)    &  0.67                 \\
     
\bottomrule    

\end{tabular}
\caption{\small Reduction in MRR and Hits@1 due to \textbf{adversarial additions on target triples in WN18RR}. Lower values indicate better results. First block of rows are the baseline attacks with random edits; second block is state-of-art attacks; remaining are the proposed attacks. For each block, we report the \emph{best} reduction in percentage relative to the original MRR.}
\label{tab:addition_WN18RR}
\end{table*}

\begin{table*}
\centering
\footnotesize
\setlength{\tabcolsep}{3.5pt}
\begin{tabular}{c  l  ll  ll   ll  ll }

\toprule

     & & \multicolumn{2}{c}{\textbf{DistMult}} & \multicolumn{2}{c}{\textbf{ComplEx}} & \multicolumn{2}{c}{\textbf{ConvE}} & \multicolumn{2}{c}{\textbf{TransE}} \\
   \cmidrule(lr){3-4}  \cmidrule(lr){5-6}  \cmidrule(lr){7-8} \cmidrule(lr){9-10} 
    & & \textbf{MRR}   & \textbf{Hits@1}  & \textbf{MRR}   & \textbf{Hits@1} & \textbf{MRR}   & \textbf{Hits@1} & \textbf{MRR}   & \textbf{Hits@1} \\
\midrule
    
     \textbf{Original}  &    & 1.00            &  1.00     &  1.00             & 1.00      &  1.00          &   1.00    & 1.00           &  1.00 \\
\midrule
    \multirow{5}{*}{\shortstack[l]{\textbf{Baseline} \\ \textbf{Attacks}}}
    & Random\_n              &  0.65 (-34\%)  & 0.50      &  0.69              &  0.57          &  0.61 (-39\%)   & 0.46        &  0.74            &  0.62      \\
    & Random\_g              &  0.66         &  0.52       &  0.66 (-34\%)     &  0.52         &  0.63            &  0.50       &   0.73 (-27\%)          &  0.61      \\
\cline{2-10} \\[-7pt]
    & Direct-Add             &  0.64 (-36\%)    &   0.48     &  0.66 (-34\%)              &  0.52     &  0.60 (-40\%)           &  0.45       &  0.72 (-28\%)           &   0.59      \\
    & CRIAGE                 &  0.66     &  0.50      &  -               &   -       &     Er         &  Er       &     -         &    -     \\
\midrule
    \multirow{9}{*}{\shortstack[l]{\textbf{Proposed} \\ \textbf{Attacks}}}
    & Dot Metric               & 0.67            & 0.54        &  0.65        & 0.50        &  0.61          &  0.46      &   0.74 (-26\%)       &  0.62       \\
    & $\bm{\ell_2}$ Metric     & 0.64            & 0.50        &  0.66        &  0.52       &  0.59 (-41\%)  &  0.43       &  0.74 (-26\%)       & 0.62         \\
    & Cos Metric               & 0.63 (-37\%)    & 0.49        & 0.63 (-37\%) & 0.47        &  0.60          &  0.43       &   0.74 (-26\%)      & 0.61         \\
\cline{2-10} \\[-7pt] 
    & GD (dot)           &  0.61 \textbf{(-39\%)}    & 0.45         &  0.65                   &  0.50       & 0.62           &  0.46       &  0.71 \textbf{(-29\%)}   &  0.58    \\
    & GL ($\ell_2$)      &  0.63                     & 0.48         &  0.67                   &   0.53      &  0.61 (-39\%)  &  0.45       &   0.74                   &  0.60     \\
    & GC (cos)           &  0.62                     &  0.46        &  0.64 \textbf{(-36\%)}  &  0.49       &  0.61 (-39\%)  &  0.45       &   0.71 \textbf{(-29\%)}  &   0.56     \\
\cline{2-10} \\[-7pt]
    
    & IF       &  0.61 \textbf{(-39\%)}         &  0.45      &   0.65 (-35\%)             & 0.50        &  0.58 \textbf{(-42\%)}     & 0.42     &  0.71 \textbf{(-29\%)}  &   0.58    \\
     
\bottomrule    

\end{tabular}
\caption{\small Reduction in MRR and Hits@1 due to \textbf{adversarial additions on target triples in FB15k-237}. Lower values indicate better results; best results for each model are in bold. First block of rows are the baseline attacks with random edits; second block is state-of-art attacks; remaining are the proposed attacks. For each block, we report the \emph{best} reduction in percentage relative to the original MRR; computed as $(poisoned - original)/original*100$.}
\label{tab:addition_FB15k-237}
\end{table*}


\section{Related Work}
\citet{cai2018comprehensive} and \citet{nickel2015review} provide a comprehensive survey of KGE models. We use the most popular models DistMult \citep{yang2015distmult}, ComplEx \citep{trouillon2016complex}, ConvE \citep{dettmers2018conve} and TransE \citep{bordes2013transe}. 

Our work is most closely related to CRIAGE \citep{pezeshkpour2019criage} and Direct Attack \citep{zhang2019kgeattack}, that study both adversarial additions and deletions against KGE models. But CRIAGE is only applicable to multiplicative models and our experiments (Section \ref{sec:evaluation}) show that Direct Attack is effective (with respect to random baselines) on additive models only. On the other hand, our instance attribution methods work for all KGE models. Recently, \citet{lawrence2021gradientrollback} propose Gradient Rollback to estimate the influence of training triples on the KGE model predictions. The original study uses the influential triples for post-hoc explanations, but they can also be used for adversarial deletions. However, the attack stores the model parameter updates for \emph{all} training triples which are in the order of millions for benchmark datasets; and our experiments (Section \ref{sec:evaluation}) show that it performs similar to random deletions. Whereas, our influence estimation methods do not require additional storage and are consistently better than random baselines on all KGE models.

We also study data poisoning attacks against KGE models in \citet{bhardwaj2021relationinferencepatterns}. Here, we exploit the inductive abilities of KGE models to select adversarial additions that improve the predictive performance of the model on a set of decoy triples; which in turn degrades the performance on target triples. These inference patterns based attacks cannot be used for adversarial deletions, but we will perform detailed comparison for adversarial additions in future. 
In parallel work, \citet{banerjee2021stealthypoisoningKGE} study risk aware adversarial attacks with the aim of reducing the exposure risk of an adversarial attack instead of improving the attack effectiveness. Also, previous studies by \citet{minervini2017adversarialsets} and \citet{cai2018kbgan} use adversarial regularization on the training loss of KGE models to improve predictive performance. But these adversarial samples are not in the input domain and aim to improve instead of degrade model performance.
Poisoning attacks have also been studied against models for undirected and single relational graph data ~\citep{zugner2018nettack, dai2018adversarialgcn,xu2020advgraphsurvey}. But they cannot be applied directly to KGE models because they require gradients of a dense adjacency matrix.

Other related work towards understanding KGE models are \citet{zhang2019interaction} and \citet{nandwani2020oxkbc} that generate post-hoc explanations in the form of sub-graphs. Also,  
\citet{trouillon2019inductive} study the inductive abilities of KGE models as binary relation properties for controlled inference tasks with synthetic datasets. Recently, \citet{allen2021interpreting} interpret the structure of KGE by drawing comparison with word embeddings.

The instance attribution methods we use are also used for post-hoc example-based explanations of black-box models \citep{molnar2019interpretablemlbook}. \citet{hanawa2021evaluationsimilaritymetrics, charpiat2019inputsimilarity, pruthi2020trackin} use Instance or Gradient Similarity on image data. Similar to us, \citet{han2020influencenlp, han2020influenceveiledtoxicity, pezeshkpour2021instanceattribution} use different instance attribution methods, but to provide post-hoc explanations on natural language.


\section{Conclusion}
We propose data poisoning attacks against KGE models using instance attribution methods and demonstrate that the proposed attacks outperform the state-of-art attacks. 
We observe that the attacks are particularly effective when the KGE model relies on few training instances to make predictions, i.e. when the input graph is sparse. 

We also observe that shallow neural architectures like DistMult, ComplEx and TransE are vulnerable to naive attacks based on Instance Similarity. 
These models have shown competitive predictive performance by proper hyperparameter tuning \citep{ruffinelli2020olddognewtricks, kadlec2017kgebaselines}, making them promising candidates for use in production pipelines. But our research shows that these performance gains can be brittle. This calls for improved KGE model evaluation that accounts for adversarial robustness in addition to predictive performance. 

Additionally, as in \citet{bhardwaj2020towardsrobustKGE, bhardwaj2021relationinferencepatterns}, we call for future proposals to defend against the security vulnerabilities of KGE models. Some promising directions might be to use adversarial training techniques or train ensembles of models over subsets of training data to prevent the model predictions being influenced by a few triples only. Specification of the model failure modes through adversarial robustness certificates will also improve the usability of KGE models in high-stake domains like healthcare and finance.


\section*{Acknowledgements}
This research was conducted with the financial support of Accenture Labs and Science Foundation Ireland (SFI) at the ADAPT SFI Research Centre at Trinity College Dublin.  The ADAPT SFI Centre for Digital Content Technology is funded by Science Foundation Ireland through the SFI Research Centres Programme and is co-funded under the European Regional Development Fund (ERDF) through Grant No. 13/RC/2106\_P2.


\section*{Broader Impact}
We study the problem of generating data poisoning attacks against KGE models. 
These models drive many enterprise products ranging from search engines (Google, Microsoft) to social networks (Facebook) to e-commerce (eBay) \citep{noy2019knowledgegraphs}, 
and are increasingly used in domains with high stakes like healthcare and finance \citep{hogan2020knowledgegraphs, bendsten2019astrazeneca}.
Thus, it is important to identify the security vulnerabilities of these models that might be exploited by malicious actors to manipulate the predictions of the model and cause system failure.
By highlighting these security vulnerabilities of KGE models, we provide an opportunity to fix them and protect stakeholders from harm. This honours the ACM Code of Ethics to contribute to societal well-being and avoid harm due to computing systems.

Furthermore, to study data poisoning attacks against KGE models, we use the Instance Attribution Methods from Interpretable Machine Learning. These methods can also be used to provide post-hoc explanations for KGE models and thus, improve our understanding of the predictions made by the models. In addition to understanding model predictions, instance based attribution methods can help guide design decisions during KGE model training. There are a vast number of KGE model architectures, training strategies and loss functions, and empirically quantifying the impact of the design choices is often challenging \citep{ruffinelli2020olddognewtricks}. Thus, we would encourage further research on exploring the use of instance attribution methods to understand the impact of these choices on the KGE model predictions. By tracing back the model predictions to the input knowledge graph, we can gain a better understanding of the success or failure of different design choices.

\bibliography{custom}
\bibliographystyle{acl_natbib}

\newpage
\appendix
\section*{Appendix}

\section{Dataset Details}
\label{apx:dataset_details}
We evaluate the proposed attacks on four state-of-art KGE models - DistMult, ComplEx, ConvE and TransE; on two publicly available benchmark datasets for link prediction\footnote{https://github.com/TimDettmers/ConvE}- WN18RR and FB15k-237. 
For the KGE model evaluation protocol, we filter out triples from the validation and test set that contain unseen entities.

To assess the attack effectiveness in \emph{degrading} performance on triples predicted as True, we need to select a set of triples that are predicted as True by the victim model.
Thus, we select a subset of the benchmark test set that has been ranked the best (i.e. ranks=1) by the victim KGE model. If this subset has more than 100 triples, we randomly sample 100 triples as the \emph{target triples}; otherwise we use all triples as target triples. We do this pre-processing step to avoid the expensive Hessian inverse computation in the Influence Functions (IF) for a large number of target triples - for each target triple, estimating the Hessian inverse (as an HVP) using the LissA algorithm requires one training epoch.

\begin{table}[h]
\centering
\footnotesize
\setlength{\tabcolsep}{5pt}
\begin{tabular}{c  l ll}
    \toprule           
    \multicolumn{2}{l}{} & \textbf{WN18RR} &  \textbf{FB15k-237} \\ 
    \midrule
    \multicolumn{2}{l}{Entities}                  &  40,559   & 14,505 \\ 
    \multicolumn{2}{l}{Relations}                 &  11       & 237 \\ 
    \multicolumn{2}{l}{Training}                  & 86,835    &  272,115            \\ 
    \multicolumn{2}{l}{Validation}                &  2,824    &  17,526   \\ 
    \multicolumn{2}{l}{Test}                      &   2,924   &  20,438    \\ 
    \midrule
    \multirow{4}{*}{\shortstack[l]{Subset \\ with \\ Best Ranks} }
    & DistMult  &   1,109   &  1,183    \\
    & ComplEx   & 1,198  &  1,238    \\
    & ConvE     &    1,106   &  901    \\
    & TransE    &    15   &   1223   \\
    \bottomrule
\end{tabular}
\caption{Statistics for WN18RR and FB15k-237.
We removed triples from the validation and test set that contained unseen entities to ensure that we do not add new entities as adversarial edits. 
The numbers above (including the number of entities) reflect this filtering.}
\label{tab:data}
\end{table}

Table \ref{tab:data} shows the dataset statistics and the number of triples which are ranked best by the different KGE models. 

\section{Training Details}
\label{apx:training_details}
\subsection{Training KGE models} 
\begin{table}[]
    \centering
    \small
    \begin{tabular}{c  c c   c c}
    \toprule
                  &  \multicolumn{2}{c}{\textbf{WN18RR}}       &   \multicolumn{2}{c}{\textbf{FB15k-237}}   \\
                  \cmidrule(lr){2-3}  \cmidrule(lr){4-5}  
                  &  \textbf{MRR}   &  \textbf{Hits@1}    &   \textbf{MRR}   &  \textbf{Hits@1}  \\
    \midrule
         DistMult &     0.48   &   0.44    &  0.34  &  0.24  \\
         ComplEx  &     0.51   &   0.47    &  0.34  &  0.25   \\
         ConvE    &     0.44   &   0.41    &  0.32  &  0.23  \\
         TransE   &     0.21   &   0.02    &  0.33  &  0.24  \\
    \bottomrule
    \end{tabular}
    \caption{MRR and Hits@1 results for original KGE models on WN18RR and FB15k-237}
    \label{tab:original_mrr}
\end{table}

We implement four KGE models - DistMult, ComplEx, ConvE and TransE. 
We use the 1-N training strategy proposed in \citet{lacroix2018canonical} but we do not add the reciprocal relations.
Thus, for each triple, we generate scores for $(s,r) \rightarrow o$ and $(o,r) \rightarrow s$.

For TransE scoring function, we use the L2 norm. The loss function used for all models is Pytorch's $\mathtt{ CrossEntropyLoss}$. For regularization, we use N3 regularization and input dropout on DistMult and ComplEx; input dropout, hidden dropout and feature dropout on ConvE; and L2 regularization \citep{bordes2013transe} and input dropout for TransE. 

We do not use early stopping to ensure same hyperparameters for original and poisoned KGE models. We use an embedding size of 200 for all models on both datasets. An exception is TransE model for WN18RR, where we used embedding dim = 100 due to the expensive time and space complexity of 1-N training for TransE.
We manually tuned the hyperparameters for KGE models based on suggestions from state-of-art implementations \citep{ruffinelli2020olddognewtricks, dettmers2018conve, lacroix2018canonical, ampligraph}.

Table \ref{tab:original_mrr} shows the MRR and Hits@1 for the original KGE models on WN18RR and FB15k-237. To re-train the KGE model on poisoned dataset, we use the same hyperparameters as the original model.
We run all model training, adversarial attacks and evaluation on a shared HPC cluster with Nvidia RTX 2080ti, Tesla K40 and V100 GPUs. 

To ensure reproducibility, our source code is publicly available on GitHub at \url{https://github.com/PeruBhardwaj/AttributionAttack}. The results in Section \ref{sec:evaluation} can be reproduced by passing the argument $\mathtt{reproduce-results}$ to the attack scripts. Example commands for this are available in the bash scripts in our codebase. The hyperparameter used to generate the results can be inspected in the $\mathtt{set\_hyperparams()}$ function in the file $\mathtt{utils.py}$ or in the log files.

For the LissA algorithm used to estimate the Hessian inverse in Influence Functions, we select the hyperparameter values using suggestions from \citet{koh2017understanding}. The values are selected to ensure that the Taylor expansion in the estimator converges. These hyperparameter values for our experiments are available in the function $\mathtt{set\_if\_params()}$ in the file $\mathtt{utils.py}$ of the accompanying codebase.

\subsection{Baseline Implementation Details}
One of the baselines in Section \ref{sec:evaluation} of the main paper is the Direct-Del and Direct-Add attack from \cite{zhang2019kgeattack}. The original study evaluated the method for the neighbourhood of subject of the target triple. We extend it for both subject and object to ensure fair comparison with other attacks. Since no public implementation is available, we implement our own.

\begin{table}[h]
    \centering
    \small
    \begin{tabular}{c  ccc   }
    \toprule
                  &  \multicolumn{3}{c}{\textbf{WN18RR}}         \\
                  \cmidrule(lr){2-4} 
                  &  \textbf{Original}   &  \textbf{High}    &   \textbf{Low}     \\
    \midrule
         DistMult &     1.00      &   0.98        &   0.98        \\
         ComplEx  &     1.00      &   0.96        &   0.95       \\
         ConvE    &     1.00      &   0.99       &   0.99       \\
         TransE   &     1.00      &   0.81        &   0.86        \\
         
    \midrule
                 &   \multicolumn{3}{c}{\textbf{FB15k-237}}   \\
                  \cmidrule(lr){2-4}
                  &  \textbf{Original}   &  \textbf{High}    &   \textbf{Low} \\
    \midrule
        DistMult    &  1.00     &  0.64    &   0.64     \\
        ComplEx     &  1.00     &  0.67    &   0.66     \\
        ConvE       &  1.00     &  0.62     &   0.60    \\
        TransE      &  1.00     &  0.72    &   0.73     \\
                  
    \bottomrule
    \end{tabular}
    \caption{MRR of KGE models trained on original datasets and poisoned datasets from the Direct-Add baseline attack in \citet{zhang2019kgeattack}. High, Low indicate the high (20\%) and low percentage (5\%) of candidates selected from random down-sampling. }
    \label{tab:ijcai_results}
\end{table}

The Direct-Add attack is based on computing a perturbation score for all possible candidate additions. Since the search space for candidate additions is of the order $\mathcal{E} \times \mathcal{R}$ (where $\mathcal{E}$ and $\mathcal{R}$ are the set of entities and relations), it uses random down sampling to filter out the candidates. The percent of triples down sampled are not reported in the original paper and a public implementation is not available. So, in this paper, we pick a high and a low value for the percentage of triples to be down-sampled and generate adversarial additions for both fractions. We arbitrarily choose 20\% of all candidate additions for high; and 5\% of all candidate additions as low.

Thus, we generate two poisoned datasets from the attack - one that used a high number of candidates and another that used a low number of candidates. We train two separate KGE models on these datasets to assess the baseline performance. Table \ref{tab:ijcai_results} shows the MRR of the original model; and poisoned KGE models from attack with high and low down-sampling percents. The results reported for Direct-Add in Section \ref{sec:evaluation} of the main paper are the better of the two results (which show more degradation in performance) for each combination.

\section{Further Analysis of Proposed Attacks}
\subsection{Runtime Analysis}
\label{apx:runtime_analysis}

We analyze the runtime efficiency of baseline and proposed attack methods for adversarial deletions. For brevity, we consider the attacks on DistMult model, but the results on other models show similar time scales. Table \ref{tab:runtime} shows the time taken in seconds to select the influential triples for DistMult model on WN18RR and FB15k-237.

\begin{table}[h]
\centering
\small
\setlength{\tabcolsep}{3.5pt}
\begin{tabular}{c  l  ll  }

\toprule
     & & \textbf{WN18RR} & \textbf{FB15k-237} \\
   
\midrule
    \multirow{6}{*}{\shortstack[l]{\textbf{Baseline} \\ \textbf{Attacks}}}
    & Random\_n              &  0.024    & 0.057         \\
    & Random\_g              &  0.002     &  0.002        \\
\cline{2-4} \\[-7pt]
    & Direct-Del            &  0.407     &  0.272          \\
    & CRIAGE                 &  2.235     &  75.117       \\
    & GR     & 29.919     &  174.191        \\
\midrule
    \multirow{7}{*}{\shortstack[l]{\textbf{Proposed} \\ \textbf{Attacks}}}
    & Dot Metric               & 0.288            & 0.342              \\
    & $\bm{\ell_2}$ Metric     & 0.057            & 0.067               \\
    & Cos Metric               & 0.067            & 0.148        \\
\cline{2-4} \\[-7pt] 
    & GD (dot)           & 7.354           & 109.015               \\
    & GL ($\bm{\ell_2}$)           & 8.100           & 120.659                      \\
    & GC (cos)           &  9.478           &  141.276                    \\
\cline{2-4} \\[-7pt]
    
    & IF       &  4751.987           &  4750.404                 \\
     
\bottomrule    

\end{tabular}
\caption{Time taken in seconds for baseline and proposed attacks to generate influential triples for DistMult on WN18RR and FB15k-237}
\label{tab:runtime}
\end{table}

We see that the Instance Similarity metrics (dot metric, $\bm \ell_2$ metric, cos metric) are more efficient than the state-of-art attacks (Direct-Del, CRIAGE and GR). Furthermore, the $\bm \ell_2$ metric is almost as quick as random triple selection.
The efficiency of Gradient Similarity metrics is also better than or equivalent to CRIAGE and GR.

Only the attack method based on IF is much slower than any other method. This is because estimating the Hessian inverse in IF requires one training epoch for every target triple, that is, we run 100 training epochs to get the influential triples for 100 target triples. However, our results in Section \ref{sec:eval_if} of the main paper show that this expensive computation does not provide improved adversarial deletions, and thus, might be unnecessary to select influential triples for KGE models.

\begin{table*}[t]
    \centering
    \footnotesize
    \begin{tabular}{l l l}
    \toprule
        \textbf{Target Relation} &  \textbf{Influential Relation} \\
        \hline
       $\mathtt{\_has\_part}$  & $\mathtt{\_has\_part}$ \\
       $\mathtt{\_synset\_domain\_topic\_of}$ & $\mathtt{\_synset\_domain\_topic\_of}$ \\
       $\mathtt{\_has\_part}$ & $\mathtt{\_has\_part}$ \\
       $\mathtt{\_synset\_domain\_topic\_of}$   & $\mathtt{\_synset\_domain\_topic\_of}$  \\
       $\mathtt{\_synset\_domain\_topic\_of}$  &  $\mathtt{\_synset\_domain\_topic\_of}$   \\
       $\mathtt{\_synset\_domain\_topic\_of}$  & $\mathtt{\_synset\_domain\_topic\_of}$  \\
       $\mathtt{\_instance\_hypernym}$ &   $\mathtt{\_instance\_hypernym}$  \\
       $\mathtt{\_synset\_domain\_topic\_of}$   &  $\mathtt{\_synset\_domain\_topic\_of}$  \\
       $\mathtt{\_instance\_hypernym}$  &  $\mathtt{\_synset\_domain\_topic\_of}$ \\
       $\mathtt{\_synset\_domain\_topic\_of}$  &   $\mathtt{\_synset\_domain\_topic\_of}$ \\
       $\mathtt{\_member\_meronym}$  &  $\mathtt{\_derivationally\_related\_form}$   \\
       $\mathtt{\_synset\_domain\_topic\_of}$ &  $\mathtt{\_synset\_domain\_topic\_of}$  \\
       $\mathtt{\_has\_part}$  &  $\mathtt{\_has\_part}$  \\
       $\mathtt{\_member\_meronym}$   & $\mathtt{\_member\_meronym}$   \\
       $\mathtt{\_synset\_domain\_topic\_of}$   &  $\mathtt{\_synset\_domain\_topic\_of}$ \\
      \bottomrule
    \end{tabular}
    \caption{Relations from the target triples and influential triples (adversarial deletions) for the cos metric on WN18RR-TransE. This combination has 15 target triples and the table shows the relations for all of them.}
    \label{tab:wn18rr_transe}
\end{table*}

\subsection{Additional Comparison with CRIAGE}
\label{apx:criage_bce}
The baseline attack method CRIAGE estimates the influence of a training triple using the BCE loss and is thus likely to be effective only for KGE models that are trained with BCE loss. In Section \ref{sec:eval_baseline}, we found that the proposed attacks are more effective than the baseline attack.

But since our original models are trained with cross-entropy loss, we perform an additional analysis of the Instance Similarity attacks against CRIAGE for the DistMult model trained with BCE loss. Table \ref{tab:bce_analysis} shows the reduction in MRR and Hits@1 due to adversarial deletions in this training setting. We find that the Instance Similarity attacks outperform the baseline for this setting as well.

\begin{table}[h]
    \centering
    \small
    \begin{tabular}{c  c c   c c}
    \toprule
                  &  \multicolumn{2}{c}{\textbf{WN18RR}}       &   \multicolumn{2}{c}{\textbf{FB15k-237}}   \\
                  \cmidrule(lr){2-3}  \cmidrule(lr){4-5}  
                  &  \textbf{MRR}   &  \textbf{Hits@1}    &   \textbf{MRR}   &  \textbf{Hits@1}  \\
    \midrule
         Original &     1.00   &   1.00    &  1.00  &  1.00  \\
         CRIAGE  &     0.67   &   0.63    &  0.63  &  0.46   \\
         Dot Metric    &     0.86   &   0.81    &  0.61  &  0.44  \\
         $\bm{\ell_2}$ Metric   &     \textbf{0.12}   &   \textbf{0.06}    &  0.60  &  0.43  \\
         Cos Metric   &     \textbf{0.12}   &   \textbf{0.06}    &  \textbf{0.58}  &  \textbf{0.38}  \\
    \bottomrule
    \end{tabular}
    \caption{Reduction MRR and Hits@1 due to adversarial deletions for DistMult (trained with BCE loss) on WN18RR and FB15k-237}
    \label{tab:bce_analysis}
\end{table}

\vfill \break
\subsection{Analysis of Instance Attribution Methods on WN18RR-TransE}
\label{apx:wn18rr_transe}

For the TransE model on WN18RR, we found that the instance attribution methods lead to effective adversarial deletions with respect to random baselines, but not adversarial additions (Section 4.1 of main paper).
A possible reason is based on the ability of TransE model hierarchical relations, i.e. the relations that represent a hierarchy between the subject and object entities. For example, $(s,\mathtt{\_has\_part},o)$ indicates that $s$ is the parent node for $o$ in a hierarchy.

We select the Instance Similarity method cos metric for further analysis. It performs the best of all instance attribution methods for adversarial deletions, but performs worse than random neighbourhood edits for adversarial additions. 
Table \ref{tab:wn18rr_transe} shows the relations in the target triples and the influential triples (i.e. adversarial deletions) selected by cos metric.

We see that the target triples contain mostly hierarchical relations like $\mathtt{\_synset\_domain\_topic\_of}$ and $\mathtt{\_has\_part}$. Also the cos metric identifies influential triples with same relations. And since our adversarial additions are only based on modifying the entity in the influential triple, these edits improve the hierarchy structure of the graph instead of breaking it. Thus, these edits perform well for adversarial deletions, but not for additions.

\newpage
\subsection{Neighbourhood Sparsity Comparison on WN18RR and FB15k-237}
\label{apx:neighbourhood_sparsity}

\begin{figure}[h]
    \centering
    \includegraphics[scale=1,width=0.5\textwidth]{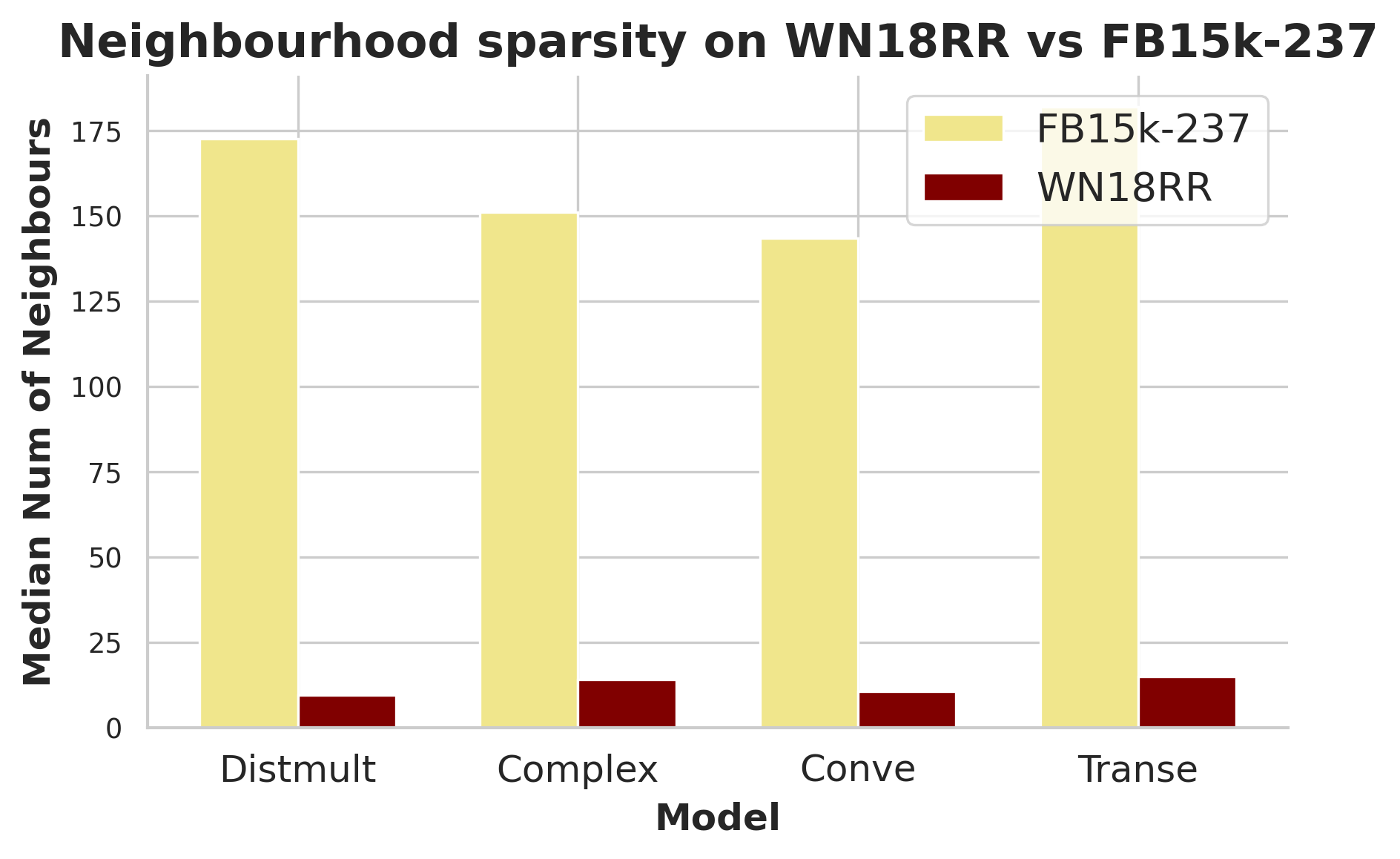}
    \caption{Comparison of the median number of neighbouring triples of target triples from WN18RR and FB15k-237 for DistMult, ComplEx, ConvE and TransE.}
    \label{fig:neighbourhood_sparsity}
\end{figure}

In Section \ref{sec:eval_data} of the main paper, we found that the proposed attacks are significantly more effective for WN18RR than for FB15k-237. 
This is likely because there are fewer triples in the neighbourhood of target triples for WN18RR than for FB15k-237.
The graph in Figure \ref{fig:neighbourhood_sparsity} shows the median number of neighbours of the target triples for WN18RR and FB15k-237. We report median (instead of mean) because of large standard deviation in the number of target triple neighbours for FB15k-237.

We see that the target triple's neighbourhood for WN18RR is significantly sparser than the neighbourhood for FB15k-237.
Thus, since the KGE model predictions are learned from fewer triples for WN18RR, it is also easier to perturb these results with fewer adversarial edits.

\end{document}